\documentclass[conference]{IEEEtran}
\IEEEoverridecommandlockouts
\usepackage{cite}
\usepackage{amsmath,amssymb,amsfonts}
\usepackage{algorithmic}
\usepackage{graphicx}
\usepackage{textcomp}
\usepackage{xcolor}
\usepackage{subcaption}
\def\BibTeX{{\rm B\kern-.05em{\sc i\kern-.025em b}\kern-.08em
    T\kern-.1667em\lower.7ex\hbox{E}\kern-.125emX}}

\usepackage[normalem]{ulem}
\newcounter{ToDo}
\newcounter{gaocomm} 
\newcounter{Note}
\renewcommand\arraystretch{1.3}
\definecolor{blue-violet}{rgb}{0.00,0.75,0.90}
\definecolor{mygreen}{rgb}{0.0, 0.5, 0.0}
\definecolor{lightblue}{rgb}{0.3,0.6,0.9}
\definecolor{awesome}{rgb}{1.0, 0.13, 0.32}
\definecolor{bostonuniversityred}{rgb}{0.8, 0.0, 0.0}

\begin{document}

\title{PREIG: Physics-informed and Reinforcement-driven Interpretable GRU for Commodity Demand Forecasting\\
}
\author{
\IEEEauthorblockN{Hongwei Ma, Junbin Gao, Minh-Ngoc Tran}
\IEEEauthorblockA{\textit{The University of Sydney }\\
Sydney, Australia \\
hongwei.ma@sydney.edu.au}

}

\maketitle

\begin{abstract}
Accurately forecasting commodity demand remains a critical challenge due to volatile market dynamics, nonlinear dependencies, and the need for economically consistent predictions. This paper introduces PREIG—a Physics-informed and Reinforcement-driven Interpretable model with GRU—a novel deep learning framework tailored for commodity demand forecasting. The model uniquely integrates a Gated Recurrent Unit (GRU) architecture with physics-informed neural network (PINN) principles by embedding a domain-specific economic constraint: the negative elasticity between price and demand. This constraint is enforced through a customized loss function that penalizes violations of the physical rule, ensuring that model predictions remain interpretable and aligned with economic theory. To further enhance predictive performance and stability, PREIG incorporates a hybrid optimization strategy that couples NAdam and L-BFGS with Population-Based Training (POP)—a reinforcement-learning inspired  mechanism that dynamically tunes hyperparameters via evolutionary exploration and exploitation. Experiments across multiple commodities datasets demonstrate that PREIG significantly outperforms traditional econometric models (ARIMA, GARCH) and deep learning baselines (BPNN,RNN) in both RMSE and MAPE. When compared with GRU, PREIG maintains good explainability while still performing well in prediction. By bridging domain knowledge, optimization theory and deep learning, PREIG provides a robust, interpretable, and scalable solution for high-dimensional nonlinear time series forecasting in economy.
\end{abstract}


\section{Introduction}

The global commodity market plays a critical role in both national and international economies. Therefore, commodity demand forecasting has become increasingly significant due to complex market dynamics, fluctuating prices, and evolving global trade policies. With commodity being the principal raw materials for industrial production, accurately predicting its demand is essential for resource planning, market stabilization, and policy-making. Recent developments in machine learning have opened up new opportunities for modeling such commodity markets. However, traditional statistical models and black-box deep learning techniques often fail to incorporate inherent physical or economic constraints that govern the underlying system dynamics.

In this context, physics-informed neural networks (PINNs) can be a promising solution. PINNs integrate data-driven modeling with embedded physical principles, ensuring that the model not only fits historical data but also adheres to known physical constraints. In the case of commodities demand forecasting, one such key physical constraint is the negative relationship between price and demand. This economic principle mandates that, all other factors being equal, the partial derivative of demand with respect to price should be negative. By embedding this constraint into the loss function of the neural network, we can enforce realistic behavior, thereby enhancing the explainability and robustness of predictions.

Our research introduces a novel forecasting framework, PREIG (Physics-informed and Reinforcement-driven Interpretable model with GRU), which incorporates a PINN structure integrated with a Gated Recurrent Unit (GRU) network. GRU, a streamlined variant of recurrent neural networks, efficiently models temporal dependencies with fewer parameters than alternatives like LSTM, thus enabling faster convergence and reduced computational complexity without sacrificing prediction accuracy \cite{XiaShaoMaSilva2021}. 
Specifically, we employ GRU as the core network structure for sequence modeling, capturing intricate time-dependent patterns inherent to commodity demand data.

The PINN part employs a composite loss function that consists of two terms. The first term is a data loss component that minimizes the error between the predicted demand and historical observations using a conventional mean squared error (MSE) loss. The second term is a physics loss component, which penalizes any violation of the physical constraint by ensuring that the partial derivative of the demand with respect to price remains negative. Formally, if $f(\mathbf{x}; \theta)$ denotes the predicted demand as a function of the input features $\mathbf{x}$ (including price $p$) and model parameters $\theta$, the physical loss is defined as:
\[
L_{\text{physics}} =\frac{1}{N} \sum_{t=1}^{N} \max\left(0, \frac{\partial f(\mathbf{x}_{1:t}; \theta)}{\partial \mathbf{p}_t}\right), 
\]
where $N$ is the whole time ponits representing the number of samples in the training dataset.

This term is incorporated into the overall loss function as
\[
L = \lambda_1 L_{\text{data}} + \lambda_2 L_{\text{physics}} 
\]
where $\lambda_1$ is a penalty hyperparameter for data loss to avoid overfitting, $\lambda_2$ is a hyperparameter that controls the trade-off between data fidelity and the enforcement of the physical constraint.

A distinct methodological advancement in our model involves an innovative hybrid optimization strategy. We combine the adaptive optimization capabilities of NAdam (Nesterov-accelerated Adaptive Moment Estimation) and the second-order approximation strengths of L-BFGS (Limited-memory Broyden–Fletcher–Goldfarb–Shanno)  with a reinforcement-learning inspired hyperparameter optimization approach, namely Population-Based Training (POP). POP dynamically evolves a population of model configurations by systematically balancing exploration (randomly perturbing less successful solutions to search new regions) and exploitation (propagating the best-performing solutions). This reinforcement-learning-driven evolution process significantly enhances the robustness and effectiveness of hyperparameter tuning, especially in complex, non-convex parameter spaces,  typically seen in deep learning \cite{JaderbergDalibardOsinderoCzarneckiDonahueRazaviVinyalsGreenDunningSimonyanFernandoKavukcuoglu2017}.  

Additionally, our framework leverages orthogonal polynomial expansion for robust denoising during preprocessing, ensuring the input data effectively represents underlying economic signals free from noisy fluctuations. 

The major contributions of our research can be summarized as follows:
\begin{enumerate}
    \item \textbf{Physics-Informed Loss Function with GRU}:
Our research integrates a novel loss function designed specifically for the GRU-based architecture, incorporating a dedicated physics loss term that enforces the economic principle of negative price elasticity. By utilizing GRU’s efficient capability in capturing temporal dependencies, our framework accurately computes the partial derivative of predicted demand with respect to price across sequential data points. This physics-informed component ensures the derivative remains consistently negative, thus strictly adhering to established economic theory. The integration of GRU not only enhances temporal predictive accuracy but also, through enforcing this domain-specific constraint, significantly improves the interpretability and robustness of the forecasting model.

\item \textbf{Hybrid Optimization Strategy Emphasizing RL-based POP}: 
We introduce an innovative parameter optimization framework that integrates the strengths of NAdam, L-BFGS, and Population-Based Training (POP). The reinforcement-learning-inspired POP methodology dynamically manages a population of candidate model configurations, systematically balancing exploration and exploitation. During optimization, POP iteratively exploits top-performing candidates by propagating their hyperparameters and simultaneously explores new solutions by introducing strategic perturbations. Following this RL-driven adaptive search process, each candidate solution is further refined through a combined NAdam and L-BFGS optimization, effectively addressing the nonconvex nature of the loss landscape. This multi-stage, RL-guided optimization strategy ensures robust convergence and superior model performance compared to conventional optimization approaches.

\item \textbf{Robust Data Preprocessing}: 
Our approach incorporates a innovative strategy to improve data quality prior to model training. We utilize a denoising step based on an orthogonal polynomial basis expansion, a sophisticated method for filtering out statistical noise from the input signals. By representing the data in a new basis of orthogonal polynomials, we can effectively isolate and remove erratic fluctuations while preserving the essential underlying trends. The direct result of this preprocessing step is a cleaner dataset that enhances the stability of the model training process, leading to more consistent performance and reliable convergence by mitigating the impact of noisy data points. 
\end{enumerate}

Our research marks a significant step forward in the field of commodity demand forecasting. We've achieved this by directly confronting the inherent limitations often found in both traditional forecasting methods and deep neural networks. Our approach isn't just about prediction; it's about delivering forecasts that are both highly accurate and profoundly interpretable. This leap is made possible through the direct integration of physical constraints into the Gated Recurrent Unit (GRU) architecture, ensuring our model adheres to economic laws. This is further bolstered by a sophisticated hybrid optimization strategy and robust preprocessing techniques, culminating in a forecasting tool that is not only exceptionally powerful but also meticulously aligned with economic theory.

\section{Literature Review}

The accurate prediction of commodity demand is crucial for effective resource management, economic stability, and strategic decision-making within the global industry\cite{CommodityMarketsOutlook}. 
This task is particularly challenging given the inherent volatility of, for example, the iron ore market, influenced by a complex interplay of macroeconomic factors, geopolitical events, technological advancements, and evolving supply chain dynamics. Early forecasting methods typically rely on traditional time series analysis techniques, such as ARIMA and GARCH models \cite{NochaiNochai2006,NadarajahAfuechetaChan2014}, which, while computationally efficient, often fall short in capturing the non-linearity and structural breaks frequently observed in iron ore market data. The underlying assumptions of stationarity and linearity inherent in these models often lead to significant forecasting errors, especially when dealing with unexpected economic shocks or geopolitical events.

Econometric models offer a more sophisticated approach by incorporating macroeconomic variables and other relevant factors to explain variations in commodities \cite{Labys2019,FattahEzzineAmanElMoussamiLachhab2018,Ampountolas2024}. 
These models attempt to establish causal relationships between commodities and variables such as global GDP growth, steel production, infrastructure investment, and government policies. However, building robust econometric models for commodity demand forecasting presents several challenges. The identification of truly relevant variables and the specification of appropriate functional forms can be highly subjective and prone to model misspecification. Furthermore, the presence of multicollinearity among predictor variables can complicate estimation and interpretation, leading to unstable and unreliable parameter estimates. The complexity of econometric models, coupled with the need for strong theoretical underpinnings, often limits their ability to adapt to changing market conditions and capture unexpected shocks.

The advent of machine learning has ushered in a new era of forecasting approaches that more effectively capture non-linear relationships and complex data patterns. Early applications of machine learning, such as Support Vector Machines (SVMs) \cite{Amin2020} 
and Artificial Neural Networks (ANNs) \cite{PanellaBarcellonaD’Ecclesia2012}, 
demonstrated notable improvements over traditional methods by handling high-dimensional data and learning intricate relationships between input variables. However, these models are often criticized as ``black boxes'', since their lack of transparency can obscure the interpretability of the decision-making process—a key requirement for industry stakeholders.

In recent years, advanced machine learning techniques have evolved to address these limitations. For example, ensemble methods like Gradient Boosting Machines (GBMs) \cite{RaySarkar2019} 
combine multiple decision trees to create more powerful predictive frameworks. Despite their improved accuracy, GBMs still grapple with issues related to interpretability, especially when applied to complex, high-dimensional datasets. Furthermore, both GBMs and ANNs typically treat the data as a purely statistical phenomenon, often ignoring the underlying physical and economic principles that govern iron ore supply and demand dynamics.

Recent developments in deep learning have further pushed the frontier of forecasting capabilities. Complex deep neural network architectures, such as deep convolutional neural networks (CNNs) \cite{ThakerChanSonner2024} 
and recurrent neural networks (RNNs) \cite{Li2024}, 
including Long Short-Term Memory (LSTM) \cite{LyTraoreDia2021} 
networks and Gated Recurrent Units (GRUs) \cite{BenAmeurBoubakerFtitiLouhichiTissaoui2023,XiaShaoMaSilva2021,ChiuHsuChenWen2023}, 
have shown considerable promise in modeling temporal dependencies and spatial features within commodity price data. Recently, within the context of recurrent neural networks, the Gated Recurrent Unit (GRU) has emerged as a compelling alternative to traditional architectures such as LSTM. The GRU simplifies the gating mechanism used in LSTMs by combining the forget and input gates into a single update gate, significantly reducing the number of trainable parameters. This simplification results in computational efficiency, faster training convergence, and robustness in modeling long-term temporal dependencies, which are particularly advantageous for demand forecasting scenarios characterized by sequential and seasonally varying patterns. And GRU has not yet been fully explored in forecasting commodities demand. 

Recent advances have seen the incorporation of physical principles into machine learning frameworks, most notably through the development of physics-informed neural networks (PINNs). Raissi et al. \cite{RaissiPerdikarisKarniadakis2019} 
introduced the concept of PINNs to solve forward and inverse problems governed by partial differential equations, thereby embedding physical laws directly into the learning process. However, PINNs in economic forecasting remains nascent. While PINNs have shown promising in various fields, including fluid dynamics \cite{KarniadakisKevrekidisLuPerdikarisWangYang2021} 
and material science \cite{PunBatraRamprasadMishin2019}, 
it is still unexplored in forecasting commodities demand.

Optimization of complex neural networks remains a critical challenge. The Adam optimizer, introduced by Kingma and Ba \cite{KingmaBa2015}, 
and its variants such as NAdam \cite{Dozat2016}, 
have been widely adopted due to their efficiency in handling sparse gradients. However, these first-order methods sometimes suffer from local minima in highly nonconvex landscapes. Liu and Nocedal \cite{LiuNocedal1989} 
provided a comprehensive review of the L-BFGS method, which leverages second-order information to accelerate convergence. More recently, Jaderberg et al. \cite{JaderbergDalibardOsinderoCzarneckiDonahueRazaviVinyalsGreenDunningSimonyanFernandoKavukcuoglu2017} 
introduced Population-Based Training (POP), based on reinforcement learning, as a way to dynamically adjust hyperparameters and escape local optima. Despite these developments, the integration of multiple optimization techniques—especially in the presence of hard physical constraints—remains underexplored.

Beyond mere model training, rigorous data preprocessing is a critical foundation for building a forecasting model that is both reliable and trustworthy. This stage involves a series of transformations designed to clean and prepare the data for analysis. Techniques like the use of orthogonal polynomial basis functions \cite{SmithAgaianAkopian2008,BrindiseVlachos2017} are particularly effective for noise reduction, as they can approximate the underlying trend in the data while filtering out random, high-frequency fluctuations. Concurrently, feature scaling is essential for standardizing the input variables, which prevents features with larger magnitudes from disproportionately influencing the model's learning process. Together, these preparatory steps are indispensable for improving the model's predictive accuracy and ensuring its stability when encountering new, unseen data.

In summary, while there have been great advances in commodity demand forecasting using both statistical and machine learning methods, existing models suffer from two major shortcomings: (i) the lack of incorporation of fundamental economic principles—such as the negative elasticity between price and demand—which makes them unexplainable and counterintuitive, and (ii) suboptimal optimization in highly nonconvex parameter spaces. The literature thus indicates a clear need for a novel forecasting framework that fuses the predictive power of deep learning with embedded physical constraints and robust optimization techniques. Our proposed approach, which employs a physics-informed GRU network optimized via a hybrid of NAdam, L-BFGS, and Population-Based Training, directly addresses these gaps. By also integrating data denoising via orthogonal polynomial expansion, our research represents a comprehensive advancement over prior models, providing enhanced predictive accuracy and economic interpretability.

\section{ Methodology}

Our forecasting model is built upon a reinforcement-learning enhanced and physics-informed GRU network that integrates data-driven learning with economic principles—specifically, the negative relationship between commodities price and demand. The structure of the model is illustrated in Fig.~\ref{fig:enter-label-1}, and the source code of the model can be find at \cite{PREIG2025}. The model consists of four major components:

\begin{figure}
    \centering
    \includegraphics[width=1.02\linewidth]{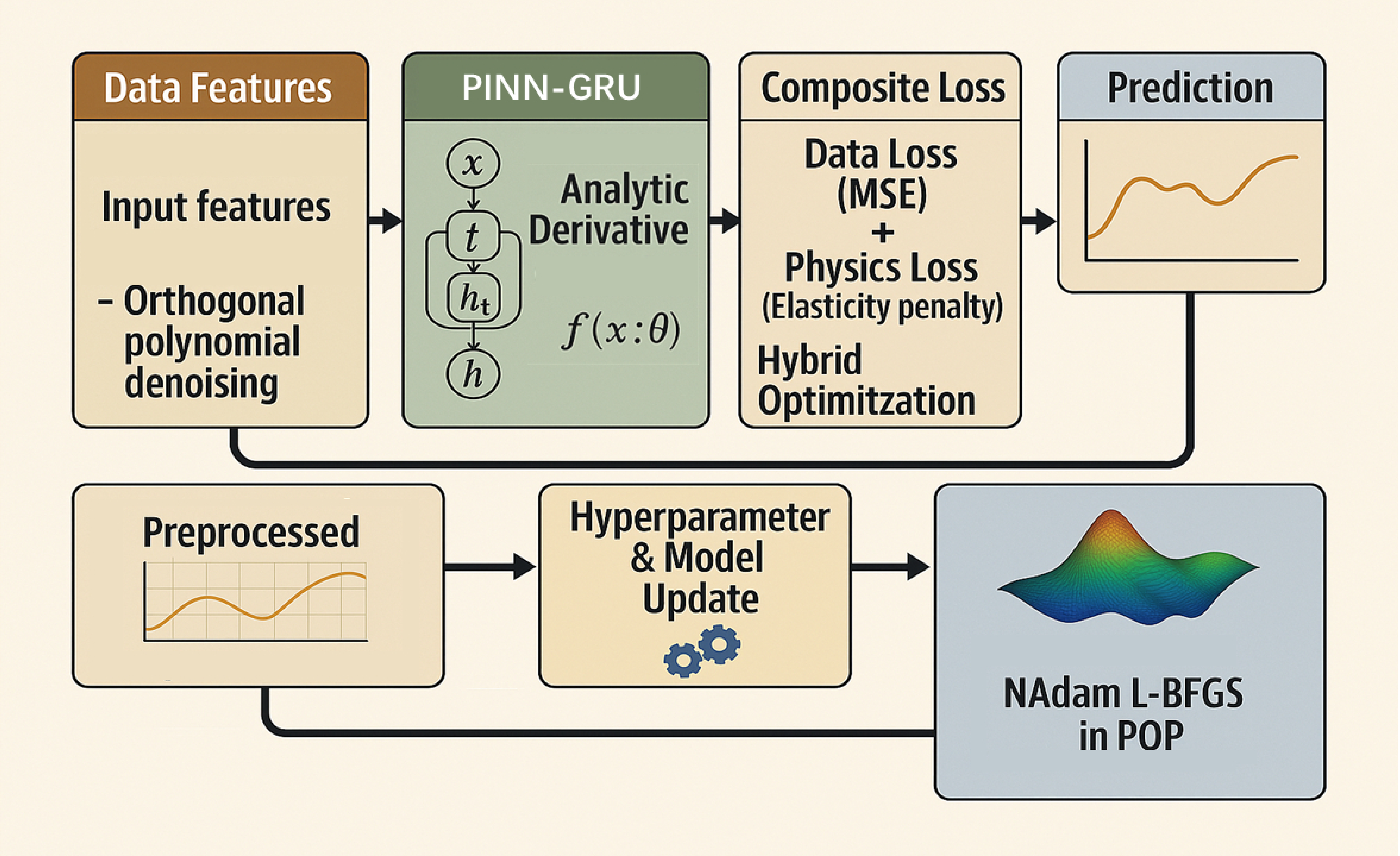}
    \caption{The Model Structure}
    \label{fig:enter-label-1}
\end{figure}

\begin{enumerate}
    \item Data Preprocessing \& Denoising via Orthogonal Polynomial Expansion
    \item GRU-based PINN Architecture \& Analytic Derivative Computation 
    \item Composite Loss Function \& Detailed Gradient Derivations
    \item Hybrid Optimization Strategy Combining POP, NAdam, and L-BFGS
\end{enumerate}

\subsection{Data Preprocessing \& Denoising via Orthogonal Polynomial Expansion  }

\subsubsection{Purpose and Rationale}
Real-world economic data are often contaminated by noise, which can obscure true patterns and lead to suboptimal model training. Our goal is to filter out high-frequency noise while preserving the dominant trends. The chosen method, orthogonal polynomial expansion, allows us to represent the input signal as a linear combination of basis functions that are statistically uncorrelated, ensuring that each basis function captures unique aspects of the data variation.  

\subsubsection{Mathematical Formulation}

\begin{enumerate}
    \item[I)] Polynomials Selection: Chebyshev Polynomials is suitable for trend component extraction in time series and performs better in minimizing the maximum error. Chebyshev Polynomials are divided into two types. The commonly used one is the first kind of Chebyshev Polynomial \(T_n(v)\), which is orthogonal on the interval $[-1, 1]$  with the weight function \(w(v)=\frac{1}{\sqrt{1-v^2}}\),satisfying:
\[\int_{-1}^1\frac{T_m(v)T_n(v)}{\sqrt{1-v^2}}dv=\frac{\pi}{2}\delta_{mn}\quad(m,n\neq0).\]
The Chebushev polynomials are defined   the recurrence relation: 
\begin{align*}
    &T_0(v) \equiv 1,\quad\quad T_1(v) = v,\\
    &T_{n+1}(v) = 2vT_n(v) - T_{n-1}(v), \text{ for }n = 2, 3, ...
\end{align*}
where \(v\) is independent variable over $[-1, 1]$.  

\item[II)] Denoising Process:  Consider the noisy observations of a particular features in data at $N$ time points, denoted by \(\mathbf{u} = [u_1, u_2, \dots, u_N]^\top\). 
We wish to denoise the signal by using a Chebyshev polynomial series up to order $K$ at $N$ locations $\{v_1, v_2, ..., v_N\}\subset [-1, 1]$ (for example, they are uniformly distributed on $[-1, 1]$), that is
\begin{align}
u_t = \sum^K_{k=0}c_k T_k(v_t) + e_t,     \quad\quad t = 1, 2, ..., N.  \label{Eq:1} 
\end{align}
Next, we construct the design matrix \(\Phi \in \mathbb{R}^{N\times (K+1)}\) as follows:  
\[  
\Phi_{t,i} = P_i(v_t),\quad \text{for } i=0,\dots,K \text{ and } t=1,\dots,N.  
\]  

Then the model \eqref{Eq:1} can be written into its matrix form:
\[  
\mathbf{u} = \Phi\,\mathbf{c} + \mathbf{e},  
\]  
where $\mathbf{c} = [c_0, c_1, ..., c_k]^\top$ and $\mathbf{e}$ is the error vector.

The objective is to minimize the least-squares error:  
\[  
\min_{\mathbf{c}} \; \|\mathbf{u} - \Phi\,\mathbf{c}\|_2^2.  
\]  
The minimizer is given by the normal equations:  
\[  
\mathbf{c} = (\Phi^\top \Phi)^{-1}\,\Phi^\top \mathbf{u}.  
\]  
The denoised signal is then reconstructed as:  
\[  
\tilde{\mathbf{u}} = \Phi\,\mathbf{c}.  
\]    

To avoid overfitting (high order) or underfitting (low order), we use cross-validation, information criteria (AIC/BIC), or residual analysis to select the optimal order $K$.  

\end{enumerate}

This initial denoising step is of critical importance. By effectively reducing the random variability and statistical dispersion within the input features, it ensures the subsequent neural network training is more stable and efficient. This refinement allows the model to dedicate its learning capacity to capturing the subtle but significant underlying economic signals, rather than erroneously fitting to statistical noise, which ultimately leads to a more robust and generalizable model.

\subsection{GRU-based PINN Architecture \& Analytic Derivative Computation  }

\subsubsection{Purpose and Rationale}
The PINN architecture is our primary modeling tool, designed to approximate the mapping from economic features to iron ore demand. In addition, it is structured to incorporate domain-specific physical constraints—most notably, the economic theory that demand decreases as price increases.  And unlike conventional fully-connected architectures used in PINN, our model employs a GRU layer as the core sequential model. The GRU efficiently handles temporal dependencies within the data by utilizing gating mechanisms (update and reset gates) to manage information flow across time steps. Formally, the GRU cell is defined as follows: 

\subsubsection{Network Structure}
Formally, the GRU cell is defined as follows: 
\[
\begin{aligned}&\mathbf{z}_t=\sigma(W_z\mathbf{x}_t+U_z\mathbf{h}_{t-1}+b_z),\\&\mathbf{r}_t=\sigma(W_r\mathbf{x}_t+U_r\mathbf{h}_{t-1}+b_r),\\&\widetilde{\mathbf{h}}_t=\tanh(W_h\mathbf{x}_t+U_h(\mathbf{r}_t\odot \mathbf{h}_{t-1})+b_h),\\&\mathbf{h}_t=(1-\mathbf{z}_t)\odot \mathbf{h}_{t-1}+\mathbf{z}_t\odot\widetilde{\mathbf{h}}_t,\end{aligned}
\]
where \(\mathbf{z}_t\) and \(\mathbf{r}_t\) are update and reset gates, respectively, \(\mathbf{x}_t\) is the input features at time  \(t\), \(\mathbf{h}_t\) is the hidden state, \(\widetilde{\mathbf{h}}_t\) is the candidate hidden state and \(\sigma\) denotes the sigmoid function. 

\subsubsection{Analytic Derivative Computation}
A unique aspect of our PINN is the enforcement of a physical constraint on the derivative of the output with respect to the price feature \(p\). The final output from the GRU at the time $t$ is the function $f(\mathbf{x}_{1:t}; \theta)$ where $\mathbf{x}_{1:t}$ represents the input features up to time $t$. Using the autodifferentation and following the chain rule, we can easily work out the Jacobian matrix $J(\mathbf{x}_t) = \frac{\partial f}{\partial \mathbf{x}_t} $   with respect to the model input \(\mathbf{x}_t\). 
If we denote by \(\mathbf{p}_t\) the price feature in \(\mathbf{x}_{1:t}\), then the derivative of price is:  
\[  
\frac{\partial f(\mathbf{x}_{1:t};\theta)}{\partial \mathbf{p}_t} = \left[J(\mathbf{x})\right]_{:,t}.  
\]  
This derivative is central to our model as it must be negative to comply with the economic principle of negative price elasticity. In subsequent sections, this derivative appears in our loss function to penalize any violation of the constraint.  

\subsection{Composite Loss Function and Detailed Gradient Derivations  }

\subsubsection{Purpose and Rationale}
The composite loss function is formulated to balance the fidelity of the model to historical data with the need to enforce an economic constraint. Specifically, while the data loss ensures that predictions are close to observed values, the physics loss penalizes any instance where the model violates the expected economic behavior (i.e., when \(\frac{\partial f}{\partial p}\) is positive).  

\subsubsection{Loss Function Components}
\begin{itemize}
    \item  Data Loss \(L_{\mathrm{data}}\):  
   Defined as the mean squared error (MSE) between predicted demand and observed demand:  
   \[  
   L_{\mathrm{data}}(\theta) = \frac{1}{N}\sum_{t=1}^N \bigl(f(\mathbf{x}_{1:t};\theta)-y_t\bigr)^2.  
   \]  
   where \(f(\mathbf{x}_{1:t};\theta)\) is the model prediction for input \(\mathbf{x}_t\) and  \(y_t\) is the actual observed demand.  

\item Physics Loss \(L_{\mathrm{physics}}\):
   Designed to enforce the economic constraint that demand must decrease as price increases:  
   \[  
   L_{\mathrm{physics}}(\theta) = \frac{1}{N}\sum_{t=1}^N \text{sum}\left(\max\!\Bigl(0,\,\frac{\partial f(\mathbf{x}_{1:t};\theta)}{\partial \mathbf{p}_t}\Bigr)\right).  
   \] 
   where $\text{sum}(\cdot)$ means the summation of vector components.
   This term contributes only when \(\frac{\partial f(\mathbf{x}_{1:t};\theta)}{\partial \mathbf{p}_t} > 0\). The use of \(\max(0,\cdot)\) ensures that no penalty is applied if the constraint is met.  

\item  Overall Loss:  
   The total loss is a weighted sum:  
   \[  
   L(\theta) = \lambda_1 L_{\mathrm{data}}(\theta) + \lambda_2\,L_{\mathrm{physics}}(\theta),  
   \]  
   where \(\lambda > 0\) controls the relative importance of adhering to the physics-based constraint compared to fitting the data.  
\end{itemize}

\subsubsection{Gradient Derivation}

To update the model parameters \(\theta\) during training, we compute the gradient of the loss with respect to \(\theta\).  
\begin{itemize}
    \item Gradient of Data Loss:
  By applying the chain rule:  
  \[  
  \nabla_\theta L_{\mathrm{data}} = \frac{2}{N}\sum_{i=1}^N \bigl(f(\mathbf{x}_{1:t};\theta)-y_t\bigr)\,\nabla_\theta f(\mathbf{x}_{1:t};\theta).  
  \]  
  Here, \(\nabla_\theta f(\mathbf{x}_{1:t};\theta)\) is computed through backpropagation and reflects how changes in the parameters affect the output. 
\item   Gradient of Physics Loss:  
  Introduce an indicator function:  
  \[  
  \delta_t = \mathbf{1}\left\{\frac{\partial f(\mathbf{x}_{1:t};\theta)}{\partial \mathbf{p}_t}>0\right\},  
  \]  
  which is 1 if the derivative violates the constraint and 0 otherwise. Then:  
  \[  
  \nabla_\theta L_{\mathrm{physics}} = 
  \frac{1}{N}\sum_{t=1}^N \delta_t\,\frac{\partial^2 f(\mathbf{x}_{1:t};\theta)}{\partial \mathbf{p}_t\,\partial\theta}.  
  \]  
  The mixed second derivative \(\frac{\partial^2 f}{\partial \mathbf{p}_t\,\partial\theta}\) is obtained via automatic differentiation tools available in modern deep learning frameworks. This derivative quantifies how a change in the model parameters affects the sensitivity of the prediction to the price feature.  
  \item Finally, the overall gradient used for optimization is:  
\[  
\nabla_\theta L = \lambda_1 \nabla_\theta L_{\mathrm{data}} + \lambda_2\,\nabla_\theta L_{\mathrm{physics}}.  
\]  
\end{itemize}

This composite gradient is crucial because it ensures that during the model's learning process, parameter adjustments are made in a synergistic way. Specifically, these updates work to simultaneously minimize the discrepancy between predicted and actual outcomes while rigorously enforcing the underlying principles of economic consistency.

\subsection{3.4 Hybrid Optimization Strategy Combining POP, NAdam, and L-BFGS  }

\subsubsection{Purpose and Rationale}
The optimization of our PINN is challenging due to the high dimensionality of the parameter space and the nonconvex nature of the composite loss function. We propose a hybrid optimization strategy that combines three complementary methods:  

\begin{itemize}
    \item Population-Based Training (POP):
   Population-based Optimization (POP) enhances the search for optimal parameters by maintaining a diverse population of candidate solutions rather than relying on a single trajectory. This inherent diversity is particularly advantageous because it significantly increases the algorithm's ability to escape local minima that might trap single-point methods. By doing so, POP effectively explores a broader parameter space, leading to a more robust and potentially superior global optimum.

\item  NAdam (Nesterov-accelerated Adaptive Moment Estimation):* 
   NAdam is a first-order optimizer that leverages adaptive learning rates and Nesterov momentum, allowing for fast convergence on noisy gradients.  

\item  L-BFGS (Limited-Memory Broyden–Fletcher–Goldfarb– Shanno):
   L-BFGS is a quasi-Newton method that approximates second-order curvature information, which helps in refining the parameters by making informed updates based on the local curvature of the loss surface. 
\end{itemize} 

\subsubsection{Detailed Steps}
\mbox{}
\begin{itemize}
    \item Stage 1: Population-Based Training (POP).  
We initialize a population of \(M\) candidate parameter sets \(\{(\theta^k,\eta^k,\lambda^k)\}_{k=1}^M\) where:  
- \(\theta^k\) represents the candidate parameters,  
- \(\eta^k\) is the candidate’s learning rate,  
- \(\lambda^k\) is its associated physics weight.  

The POP process involves:  
- Evaluation:  
  Each candidate \(\theta^k\) is trained for a fixed number of iterations using the inner optimizer (combining NAdam and L-BFGS). A validation loss \(L_{\mathrm{val}}(\theta^k)\) is computed.  
- Selection:
  Candidates are ranked by their validation loss. The top \(M/2\) candidates are retained.  
- Perturbation:  
  The bottom \(M/2\) candidates are replaced by cloning and perturbing the best-performing candidates:  
  \begin{align*}
 \theta_{\mathrm{new}} &= \theta_{\mathrm{best}} + \varepsilon\cdot\mathcal{N}(0,I),\quad  
  \eta_{\mathrm{new}} = \eta_{\mathrm{best}} \times U(0.8,1.2),\\
  \lambda_{\mathrm{new}} &= \lambda_{\mathrm{best}} \times U(0.8,1.2),       
  \end{align*}  
where \(\varepsilon\) is a small constant, \(\mathcal{N}(0,I)\) denotes Gaussian noise, and \(U(0.8,1.2)\) represents a uniformly sampled scaling factor.  

    \item Stage 2: NAdam Update.   
The NAdam update rules are:  
\begin{align*}
\mathbf{Given\ } &\beta_1,\beta_2,\varepsilon,\{\alpha_\tau\}_{\tau=1}^L,\ \ 
\mathbf{m}_0 = \mathbf{0},\quad \mathbf{n}_0 = \mathbf{0},\\
\mathbf{for}\ \tau &=1,\ldots,L:\\
\mathbf{g}_\tau &\leftarrow \nabla_{\theta_{\tau-1}} f_\tau(\boldsymbol{\theta}_{\tau-1}),\\
\mathbf{m}_\tau &\leftarrow \beta_1 \mathbf{m}_{\tau-1} + \left(1-\beta_1\right)\mathbf{g}_\tau,\\
\mathbf{n}_\tau &\leftarrow \beta_2 \mathbf{n}_{\tau-1} + \left(1-\beta_2\right)\mathbf{g}_\tau^2,\\
\widehat{\mathbf{n}}_\tau &\leftarrow \frac{\mathbf{n}_\tau}{1-\beta_2^{\tau}},\\
\widehat{\mathbf{m}}_\tau &\leftarrow \frac{\beta_1^{\tau+1} \mathbf{m}_\tau}{1-\beta_1^{\tau+1}} + \frac{(1-\beta_1)\mathbf{g}_\tau}{1-\beta_1^{\tau}},\\
\theta_\tau &\leftarrow \theta_{\tau-1} - \alpha_\tau \frac{\widehat{\mathbf{m}}_\tau}{\sqrt{\widehat{\mathbf{n}}_\tau} + \varepsilon}.
\end{align*}

    \item Stage 3: L-BFGS Refinement. 
After a block of NAdam iterations, we refine the parameters with L-BFGS to leverage curvature information. L-BFGS approximates the inverse Hessian \( H_\tau \) using a limited memory of past gradients and updates. The update rule is given by:
\[
\theta_{\varphi+1} = \theta_\varphi - H_\varphi \nabla_\theta L(\theta_\varphi).
\]

The Hessian approximation \( H_\varphi \) is computed using the formula:
\[
H_\varphi \approx \left( \sum_{j=0}^5 \frac{s_{\varphi-j} s_{\varphi-j}^\top}{\nabla_{\varphi-j}^\top s_{\varphi-j}} \right)^{-1},
\]
where \( s_\varphi = \theta_\varphi - \theta_{\varphi-1} \) and \( \nabla_\varphi = \nabla_\theta f(\theta_\varphi) - \nabla_\theta f(\theta_{\varphi-1}) \). This method accelerates convergence, particularly near local minima, and the quasi-Newton step refines the solution by approximating second-order information without computing the full Hessian, which would be computationally expensive.  
\end{itemize}

This section provides a truly comprehensive understanding by diving deep into our PREIG framework.We cover every stage, from the initial preprocessing of data and the intricate details of model formulation to the sophisticated techniques employed in optimization. This thorough approach doesn't just scratch the surface; it lays an exceptionally solid foundation for reliably forecasting commodity demand. The result is not only improved accuracy in our predictions but also a rigorous adherence to fundamental economic principles, ensuring our forecasts are both precise and economically coherent.

\section{Experiment Results and Analysis}
\subsection{Data Collection and Processing}

Our research employs monthly commodities export volume data spanning December 2014 to December 2024, with the period from December 2014 to June 2024 designated as the training set and July to December 2024 as the test set. Based on comprehensive analysis of factors influencing commodities export volume, we performed time lag on the features and select the time lag order with the highest correlation with commodity demand, then we developed a three-dimensional feature system. The first dimension captures macroeconomic indicators through metrics like exchange rates, commodities prices, and PPI. The second dimension incorporates mesoeconomic indicators including future prices and inventories, while the third dimension considers historical exports. Detailed feature descriptions are provided in Table~\ref{tab:feature_categories1}. The data is obtained from FRED, Investing, several national statistical bureaus, several national import and export departments, the International Energy Agency, the World Trade Organization and other institutions.


\begin{table}[h]
    \renewcommand{\arraystretch}{1.6}
    \centering
    \begin{tabular}{|p{3cm}|p{3cm}|} 
    \hline
    \textbf{Feature Category} & \textbf{Variables} \\ \hline
    & Exchange Rates \\ \cline{2-2} 
    & Commodities Prices \\ \cline{2-2} 
    Macroeconomic & PPI \\ \cline{2-2} 
    & GDP \\ \cline{2-2} 
    & M2 \\ \cline{2-2} 
    & Industry Production \\ \hline
    & Energy Price \\ \cline{2-2} 
    & Substitutes Price \\ \cline{2-2} 
    Mesoeconomic & Commodity Price \\ \cline{2-2} 
    & Future Price \\ \cline{2-2} 
    & Shipping Rates \\ \cline{2-2} 
    & Inventory \\ \hline
    Historical & Historical Exports \\ \hline
    \end{tabular}
    \caption{Feature Description}
    \label{tab:feature_categories1}
\end{table}

Due to substantial value range variations across different feature types, we normalized all input data using following Equation  prior to model training, \[\mathbf{x}^*=\displaystyle\frac{\mathbf{x}-\mathbf{x}_{\min}}{\mathbf{x}_{\max}-\mathbf{x}_{\min}}.\] 

This standardization process enhances neural network learning efficiency by ensuring comparable feature scales while maintaining their relative relationships. The methodological approach systematically addresses key forecasting challenges through its comprehensive feature selection, rigorous statistical validation, and appropriate data preprocessing, while maintaining clear temporal separation between training and evaluation periods to ensure valid performance assessment.

Our research employs the proposed loss function during training, and employs the proposed optimization method to minimize the loss function through systematic parameter tuning.  Experiments have shown that our method demonstrates efficiency in practical applications compared to other adaptive learning algorithms. 

For predictive performance evaluation, two complementary metrics are adopted: Root Mean Squared Error (RMSE) and Mean Absolute Percentage Error (MAPE). RMSE preserves the original measurement units of the target variable, offering intuitive interpretation and heightened sensitivity to significant prediction errors. MAPE provides a normalized, scale-independent assessment essential for cross-model comparisons. Their mathematical formulations are provided respectively\cite{Eğrioğlu2008},

\[\text{RMSE}=\sqrt{\frac{1}{N}\sum_{t=1}^{N}(y_t-f(\mathbf{x}_{1:t};\theta))^2},\]
\[\text{MAPE}=\frac{1}{N}\sum_{t=1}^N\left|\frac{y_t-f(\mathbf{x}_{1:t};\theta)}{y_t}\right|\times100\%.\]

RMSE's dimensional consistency makes it particularly valuable for identifying substantial prediction deviations in absolute terms, while MAPE's percentage-based measurement enables direct interpretation of relative error magnitudes across different scales. This dual-metric approach ensures comprehensive accuracy assessment by capturing both absolute and relative error dimensions.

By combining the proposed efficient optimization method with rigorous evaluation through RMSE and MAPE, the methodology establishes a robust framework for commodity demand forecasting. The approach balances computational efficiency with thorough performance validation, leveraging each technique's distinct advantages while mitigating their individual limitations through strategic integration.

\begin{figure}
  \centering
  \begin{subfigure}{0.45\textwidth}
    \includegraphics[width=\linewidth]{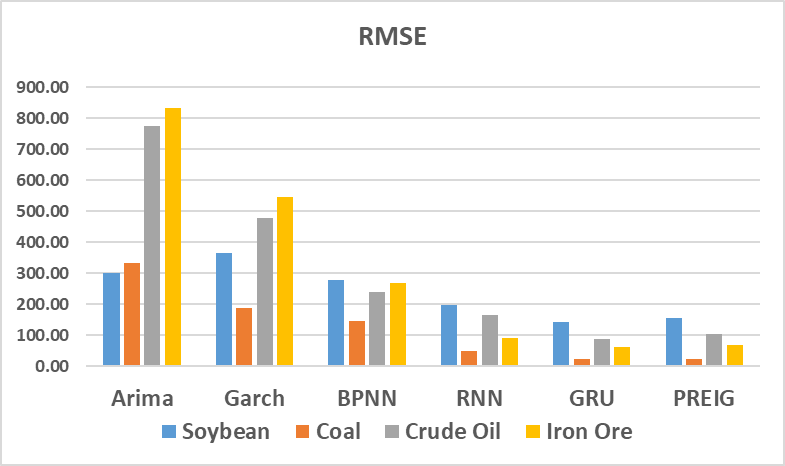}
    \caption{RMSE Comparison of Six Models}
    \label{fig2:sub-first} 
  \end{subfigure}
  \hfill
  \begin{subfigure}{0.45\textwidth}
    \includegraphics[width=\linewidth]{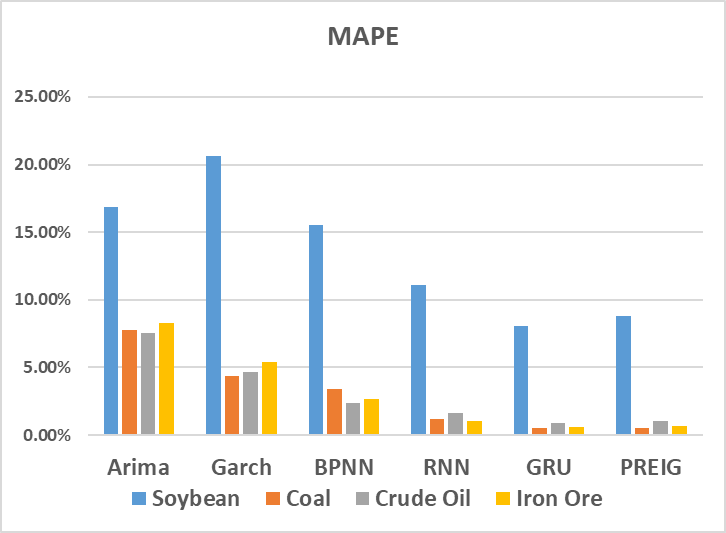}
    \caption{MAPE Comparison of Six Models}
    \label{fig2:sub-second}
  \end{subfigure}
  \caption{Comparison of model performance metrics.} 
  \label{fig:performance_comparison} 
\end{figure}

Our research compares the predictive performance of the proposed PREIG model against four established methods in commodity demand forecasting: the ARIMA time series model, Garch time series model, Backpropagation (BP) Neural Network and GRU. The experimental results of Coal dataset, presented in Table~\ref{tab:2}, demonstrate each model's forecasting accuracy. Subsequent error calculations derived from Table 2's data are shown in Table~\ref{tab:3} and Table~\ref{tab:4}, enabling direct comparison of prediction quality across methodologies. Fig~\ref{fig:performance_comparison} visualizes the RMSE and MAPE of our model and baseline models, clearly illustrating the substantial improvement our model offers over existing baseline methods. 

\begin{table*}[]
\centering
\begin{tabular}{|llllllllllllll}
\hline
       &        & \multicolumn{2}{l}{Arima} & \multicolumn{2}{l}{Garch} & \multicolumn{2}{l}{BPNN} & \multicolumn{2}{l}{RNN} & \multicolumn{2}{l}{GRU} & \multicolumn{2}{l|}{PREIG}              \\
Time   & Real   & Predicted     & Error     & Predicted     & Error     & Predicted    & Error     & Predicted    & Error    & Predicted    & Error    & Predicted & \multicolumn{1}{l|}{Error}  \\ \hline
Jul-24 & 30.083 & 27.661        & 8.05\%    & 31.440        & 4.51\%    & 29.313       & 2.56\%    & 28.78        & 4.33\%   & 30.272       & 0.63\%   & 29.881    & \multicolumn{1}{l|}{0.67\%} \\
Aug-24 & 30.394 & 27.498        & 9.53\%    & 29.300        & 3.60\%    & 33.121       & 8.97\%    & 31.38        & 3.25\%   & 30.458       & 0.21\%   & 29.647    & \multicolumn{1}{l|}{2.46\%} \\
Sep-24 & 30.614 & 33.139        & 8.25\%    & 27.813        & 9.15\%    & 30.834       & 0.72\%    & 31.47        & 2.81\%   & 29.943       & 2.19\%   & 30.066    & \multicolumn{1}{l|}{1.79\%} \\
Oct-24 & 30.868 & 33.285        & 7.83\%    & 29.794        & 3.48\%    & 29.831       & 3.36\%    & 31.01        & 0.46\%   & 30.445       & 1.37\%   & 30.873    & \multicolumn{1}{l|}{0.01\%} \\
Nov-24 & 31.250 & 33.934        & 8.59\%    & 31.216        & 0.11\%    & 32.278       & 3.29\%    & 30.76        & 1.58\%   & 31.494       & 0.78\%   & 31.789    & \multicolumn{1}{l|}{1.73\%}  \\
Dec-24 & 31.897 & 30.554        & 4.21\%    & 34.273        & 7.45\%    & 33.581       & 5.28\%    & 31.05        & 2.64\%   & 30.991       & 2.84\%   & 31.574    & \multicolumn{1}{l|}{1.01\%} \\ \hline
\end{tabular}
\caption{Experimental Prediction Results on Coal Dataset(Million Tons)}
\label{tab:2}
\end{table*}

\begin{table}[]
\centering
\begin{tabular}{|lllllll|}
\hline
          & Arima   & Garch   & BPNN    & RNN    & GRU    & PREIG  \\ \hline
Soybean   & 300.20  & 366.80  & 276.74  & 197.37 & 143.82 & 154.55 \\ \hline
Coal      & 3.34    & 1.87    & 1.46    & 0.51   & 0.24   & 0.22   \\ \hline
Crude Oil & 774.86  & 478.50  & 240.79  & 164.82 & 89.53  & 104.94 \\ \hline
Iron Ore  & 8336.04 & 5453.20 & 2701.40 & 922.49 & 627.95 & 685.23 \\ \hline
\end{tabular}
\caption{RMSE Comparison of Six Models}\label{tab:3}
\end{table}

\begin{table}[]
\centering
\begin{tabular}{|lllllll|}
\hline
          & Arima   & Garch   & BPNN    & RNN     & GRU    & PREIG \\ \hline
Soybean   & 16.89\% & 20.63\% & 15.57\% & 11.10\% & 8.09\% & 8.78\% \\ \hline
Coal      & 7.78\%  & 4.36\%  & 3.41\%  & 1.19\%  & 0.55\% & 0.52\% \\ \hline
Crude Oil & 7.53\%  & 4.65\%  & 2.34\%  & 1.60\%  & 0.87\% & 1.02\% \\ \hline
Iron Ore  & 8.27\%  & 5.41\%  & 2.68\%  & 1.05\%  & 0.62\% & 0.69\% \\ \hline
\end{tabular}
\caption{MAPE Comparison of Six Models}\label{tab:4}
\end{table}

Tables~\ref{tab:3} and \ref{tab:4} demonstrate the proposed PREIG model's superior predictive accuracy, reducing overall RMSE by 2117.375 and MAPE by 7.37\%
compared to ARIMA. When evaluated against Garch, BP Neural Network, the model achieves MAPE reductions of 6.01\% and 3.25\% respectively. When compared with GRU, PREIG maintains good explainability while still performing well in prediction, and even surpasses GRU on one dataset. These results confirm that for high-dimensional nonlinear time series forecasting, particularly in commodity demand prediction, the PREIG model delivers exceptional performance while maintaining good explainability.

\begin{figure}
    \centering
    \includegraphics[width=1.02\linewidth]{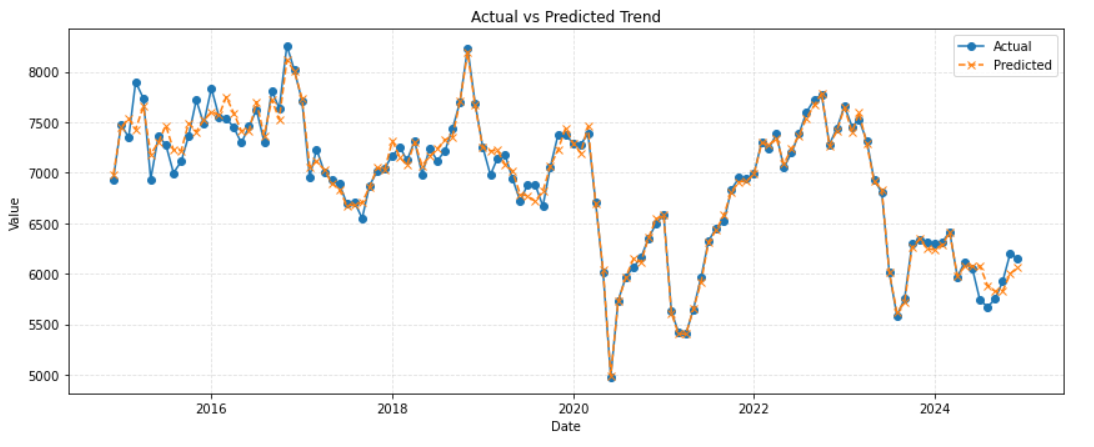}
    \caption{Fitting and Prediction Results on Crude Oil Dataset}
    \label{fig:enter-label3}
\end{figure}

\begin{figure}
    \centering
    \includegraphics[width=1.02\linewidth]{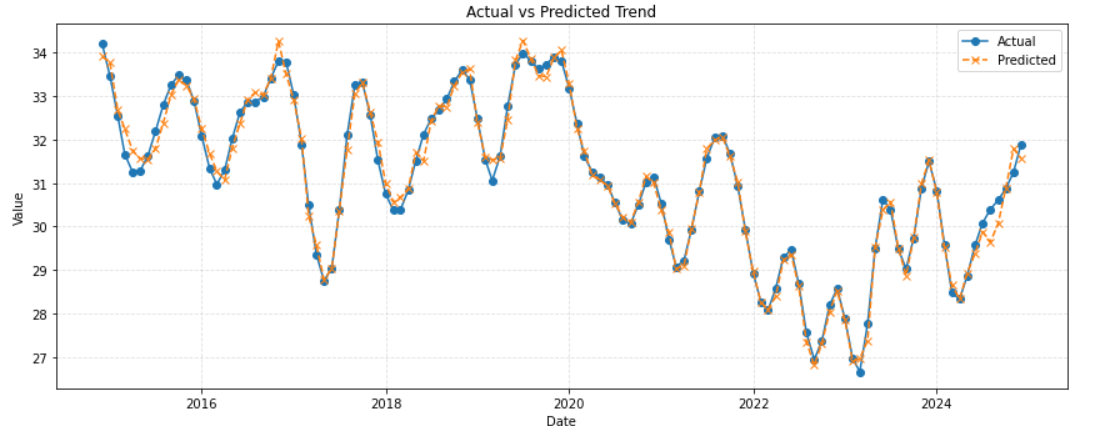}
    \caption{Fitting and Prediction Results on Coal Dataset}
    \label{fig:enter-label4}
\end{figure}

\begin{figure}
    \centering
    \includegraphics[width=1.02\linewidth]{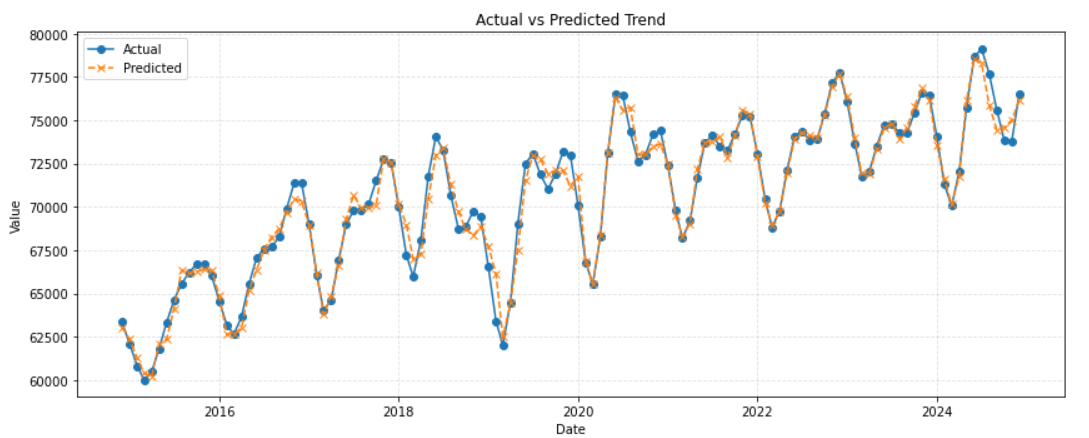}
    \caption{Fitting and Prediction Results on Iron Ore Dataset}
    \label{fig:enter-label5}
\end{figure}

\begin{figure}
    \centering
    \includegraphics[width=1.01\linewidth]{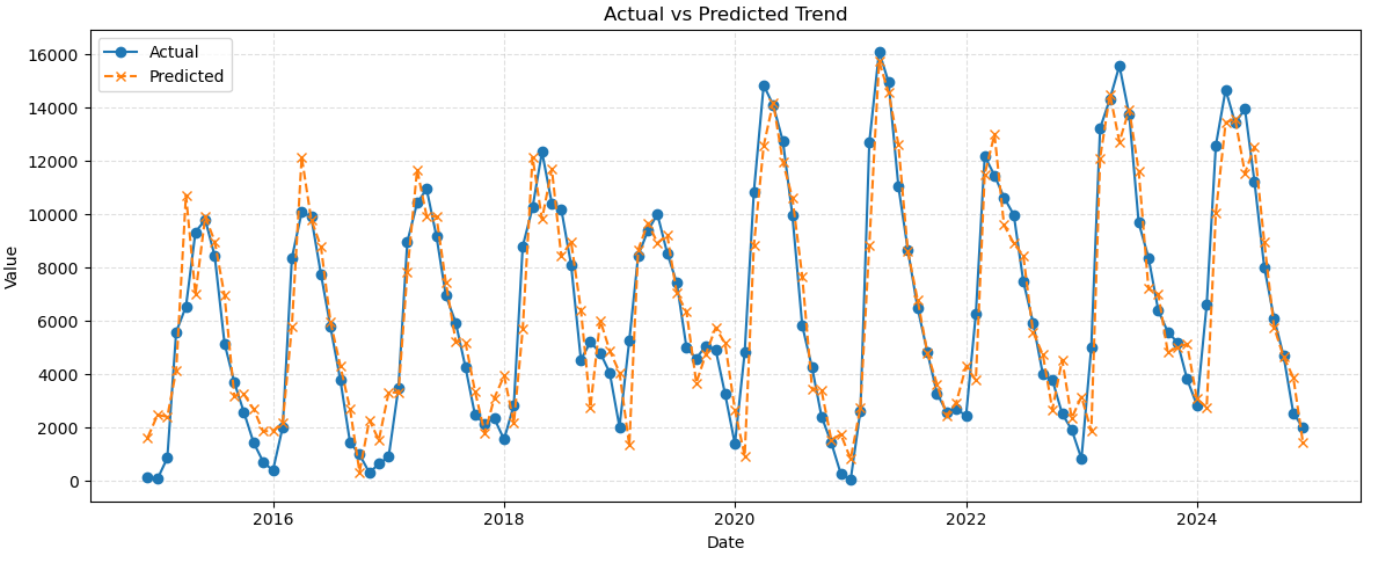}
    \caption{Fitting and Prediction Results on Soybean Dataset}
    \label{fig:enter-label6}
\end{figure}

Beyond traditional metrics, our research incorporates the trend fitting between predicted and actual values as a key evaluation indicator. Figs. \ref{fig:enter-label3}--\ref{fig:enter-label6} visually confirm the effectiveness of the PREIG model, displaying highly accurate fitting and prediction outcomes. These figures demonstrate that the trends in commodity demand predicted by the PREIG model fundamentally align with the actual demand patterns. More critically, the model exhibits remarkable stability, as evidenced by its ability to maintain predictions that do not deviate significantly from actual values even when demand fluctuates wildly. This consistency strongly validates the PREIG model's superior predictive capabilities and its reliability in dynamic market conditions.

\section{Conclusion}

Our research introduced PREIG, a physics-informed, reinforcement-driven, and GRU-based framework that couples domain knowledge with deep temporal modeling and an RL-enhanced hybrid optimizer. By embedding the negative price–demand elasticity directly into the loss function, PREIG delivers predictions that remain both accurate and economically consistent. Empirical results across four commodity datasets show that our model reduces RMSE and MAPE relative to ARIMA, GARCH, and BPNN baselines, and maintains good explainability compared with GRU . The orthogonal-polynomial denoising pipeline and the POP–NAdam–L-BFGS optimization schedule further contribute to training stability, faster convergence, and reliable generalization. Collectively, these contributions position PREIG as a practical, interpretable, and scalable alternative for high-dimensional nonlinear time series economic forecasting tasks. 
 
Several limitations warrant discussion: First, the physics-guided constraint currently implements a univariate price derivative; cross-elasticities among correlated commodities and multi-factor economic constraints remain unaddressed. Second, the hybrid optimizer incurs non-trivial computational overhead, especially during POP’s population evaluation phase, which may impede real-time deployment on large feature spaces.  

Our research contributes to the field by: 
\begin{itemize}
    \item Embedding the negative price-demand elasticity as a differentiable penalty inside a GRU, yielding predictions that remain consistent with economic theory while capturing long-range temporal structure.  \item A three-stage pipeline—POP for hyper-parameter evolution, NAdam for fast first-order updates, and L-BFGS for second-order refinement—delivers faster convergence and superior minima in non-convex landscapes. \item  Extensive tests on coal, crude oil, soybean, and iron-ore datasets showcase state-of-the-art accuracy and robustness, establishing PREIG as a strong candidate for deployment in commodity-demand forecasting pipelines. 
\end{itemize}

In conclusion, our research presents a novel and robust framework, PREIG, which effectively integrates economic principles with advanced deep learning and optimization techniques. By successfully embedding price-demand elasticity into a GRU model and employing a sophisticated hybrid optimization strategy, the study demonstrates state-of-the-art performance in commodity demand forecasting. While acknowledging current limitations, such as constraints and computational overhead, the work lays a strong foundation for future advancements. The promising results across multiple real-world datasets underscore PREIG's potential as a practical, interpretable, and scalable solution, marking a significant step forward in applying physics-informed machine learning to complex economic time series analysis.

\bibliographystyle{IEEEtran}  
\bibliography{myreference}  

\end{document}